\documentclass[runningheads]{llncs}
\usepackage[T1]{fontenc}
\usepackage{graphicx}
\usepackage{cancel}
\usepackage{hyperref}
\usepackage{bm}
\usepackage{balance}

\usepackage{amssymb}
\usepackage{mathtools}
\usepackage{amsmath}
\usepackage{cases}
\usepackage{cite}
\usepackage{multirow}
\usepackage{comment}
\usepackage{derivative}
\usepackage{mwe}
\usepackage{subcaption}
\usepackage{calc}  

\usepackage{tikz}
\usetikzlibrary{shapes.geometric, arrows}

\tikzstyle{block} = [rectangle, rounded corners, minimum width=3.5cm, minimum height=1cm,text centered, draw=black, fill=blue!20]
\tikzstyle{arrow} = [thick,->,>=stealth]

\usepackage{amsthm} 
\theoremstyle{definition}

\usepackage{xcolor}
\usepackage{eso-pic}
\AddToShipoutPictureBG*{%
  \AtPageUpperLeft{%
    \setlength{\unitlength}{1mm}%
    \put(-35.,-9){\makebox(\paperwidth,0)[c]{\parbox{0.9\textwidth}{Agents and Robots for reliable Engineered Autonomy, AREA 2025}}}
  }
}

\AddToShipoutPictureBG*{%
  \AtPageUpperLeft{%
    \setlength{\unitlength}{1mm}%
    \put(-10.5,-15){\makebox(\paperwidth,0)[c]{\parbox{1.3\textwidth}{Communications in Computer and Information Science, Springer}}}
  }
}

\begin{document}
\title{BC-MPPI: A Probabilistic Constraint Layer for Safe Model-Predictive Path-Integral Control}
\titlerunning{BC-MPPI: Probabilistic Constraints for safe MPPI} 
%
%
\author{Odichimnma Ezeji\inst{1} \and
        Michael Ziegltrum\inst{1} \and
        Giulio Turrisi\inst{2} \and
        Tommaso Belvedere\inst{3} \and
        Valerio Modugno\inst{1}}

\authorrunning{O. Ezeji et al.}   

\institute{Department of Computer Science, University College London, UK\\
           \email{\{chidimma.ezeji.21, michael.ziegltrum.24, v.modugno\}@ucl.ac.uk}
           \and
           Istituto Italiano di Tecnologia (IIT), Genova, Italy\\
           \email{giulio.turrisi@iit.it}
           \and
           CNRS, Inria, IRISA, Campus de Beaulieu, Rennes Cedex, France\\
           \email{tommaso.belvedere@irisa.fr}}
\maketitle              
\begin{abstract}
Model Predictive Path Integral (MPPI) control has recently emerged as a fast, gradient-free alternative to model-predictive control in highly non-linear robotic tasks, yet it offers no hard guarantees on constraint satisfaction.  We introduce \emph{Bayesian-Constraints MPPI} (BC-MPPI), a lightweight safety layer that attaches a probabilistic surrogate to every state and input constraint.  At each re-planning step the surrogate returns the probability that a candidate trajectory is feasible; this joint probability scales the weight given to a candidate, automatically down-weighting rollouts likely to collide or exceed limits and pushing the sampling distribution toward the safe subset; no hand-tuned penalty
costs or explicit sample rejection required.  We train the surrogate
from $1{,}000$ offline simulations and deploy the controller on a
quadrotor in MuJoCo with both static and moving obstacles.  Across
$K\!\in\![100,1500]$ rollouts BC-MPPI preserves safety margins while satisfying the prescribed probability of violation. Because the surrogate is a stand-alone, version-controlled artefact and the runtime safety score is a single scalar, the approach integrates naturally with verification-and-validation pipelines for certifiable autonomous systems.

 \and Autonomous Systems
\end{abstract}
\section{Introduction}
Model Predictive Control (MPC) was first developed in the process-control community \cite{Qin2003} and has since become a workhorse in robotics, underpinning behaviors as diverse as mobile-robot navigation, manipulation, legged locomotion, and aerial acrobatics \cite{Katayama2023}. Its appeal lies in the explicit use of a predictive model to optimize a finite-horizon cost while respecting user-defined constraints at every step. Yet this very strength is also a weakness: effective deployment still depends on hand-crafting analytic cost terms and constraint sets that are differentiable, well-conditioned, and sufficiently rich to capture task objectives and safety limits. For highly nonlinear, contact-rich, or perception-driven tasks—where objectives may be implicit in sensor data or learned from experience—defining such functions becomes a substantial engineering burden and can limit MPC’s practicality in advanced robotic applications \cite{Wensing2024}.\\

Stochastic sampling–based Model Predictive Path Integral (MPPI) control offers a compelling alternative. MPPI sidesteps gradient evaluations by estimating the path integral of cost along thousands of Monte-Carlo rollouts, allowing it to tackle highly nonlinear, non-convex dynamics and non-differentiable objectives. It has delivered state-of-the-art performance in aggressive autonomous driving \cite{Williams2016,Williams2017}, agile quadrotor flight \cite{Minarik2024, belvedere2025feedbackmppi}, and contact-rich quadruped locomotion \cite{Carius2022,Turrisi2024}. The adoption of GPU-accelerated sampling and differentiable programming frameworks now enables sub-millisecond evaluation of tens of thousands of trajectories, bringing MPPI firmly into the real-time regime \cite{Turrisi2024}. At the same time, techniques such as low pass filtering \cite{Kicki2025} and learned importance sample priors \cite{howell2022} mitigate the characteristic action noise of MPPI, reducing oscillations and improving closed-loop stability.\\

However, where MPC excels in hard constraint satisfaction MPPI struggles. Its Monte-Carlo nature makes it difficult to guarantee that every sampled trajectory respects state and input limits, and naive penalty costs often lead to brittle tuning and constraint-violation outliers. Embedding a principled constraint-handling layer within MPPI, therefore, remains an open challenge, especially for safety-critical robotic applications that must certify collision-avoidance, torque limits, or contact-stability conditions in real time.\\


In this work we close that gap with Bayesian Constraints MPPI (BC-MPPI), a safety layer that learns a probabilistic description of task constraints and folds it directly into the MPPI sampling process. We represent each hard constraint with a Bayesian surrogate - a Bayesian neural network (BNN) - which returns both a mean estimate of constraint satisfaction and an epistemic uncertainty measure \cite{Rasmussen2006,Neal1995}. At every control step, these surrogates reshape the MPPI proposal distribution: trajectories that venture into regions with high violation probability or high model uncertainty are exponentially down-weighted. At the same time, those that remain in the safe set are sampled more densely. 
We validate BC-MPPI in a high-fidelity quadrotor simulator, where the drone must fly point-to-point while respecting static obstacles (fixed walls, ceiling, and floor) and dynamic constraints (moving no-fly zones and time-varying thrust limits). The experiments show that the learned Bayesian constraint layer steers the sampling toward safe rollouts, keeping the violation probability below 
1 without sacrificing trajectory optimality.

\section{Related Works}
Early efforts to inject constraints into stochastic optimal control pre-date MPPI itself. Cross-Entropy Motion Planning (CEM) biases a Gaussian distribution over control sequences via the cross-entropy method so that almost all sampled rollouts satisfy kinodynamic and obstacle constraints, enabling agile ground-vehicle and quadrotor manoeuvres without requiring gradients \cite{Kobilarov2012}. In model-based reinforcement learning, Robust CEM Planning couples an ensemble dynamics model with a violation-budgeted CEM optimiser; uncertainty and sparse hazard rewards are folded into the cost, dramatically reducing failure counts on Safety-Gym tasks while maintaining sample efficiency \cite{Liu2021}.\\

Within MPPI, the simplest strategy is to encode safety softly into the objective. Risk-Aware MPPI (RA-MPPI) replaces the expected cost with Conditional Value-at-Risk, steering optimisation toward the worst-case tail of the rollout distribution; on autonomous-racing benchmarks, it achieves baseline lap times with an order-of-magnitude fewer crashes \cite{Yin2022}. Because constraints appear only as penalties, however, violations remain possible whenever no high-quality feasible sample is drawn.\\

A contrasting line of work enforces hard constraints through projection or real-time filtering. Constrained Stochastic Optimal Control projects every Monte-Carlo rollout onto equality manifolds and surrounds inequalities with differentiable barrier functions, guaranteeing feasibility on manipulators and legged robots while preserving MPPI’s gradient-free character \cite{Carius2022}. Shield-MPPI passes each MPPI command through a discrete-time control-barrier-function (CBF) quadratic programme, giving zero off-track events in aggressive racing on CPU-only hardware \cite{Yin2023}. Extending this idea to hybrid dynamics, Risk-Aware MPPI for Stochastic Hybrid Systems encodes timed reach-avoid objectives in the cost while a companion CBF filter ensures collision-free motion around moving obstacles with formal guarantees \cite{Parwana2024}. Projection and filtering, however, can over-constrain the optimiser and sacrifice solution optimality.\\

The closest work to ours shapes the sampling distribution itself so that feasible, low-risk rollouts are drawn more often. Robust MPPI perturbs a nominal tube of trajectories and minimises a free-energy bound to keep an AutoRally car strictly within track limits despite disturbances \cite{Gandhi2021}. Dynamic Risk-Aware MPPI computes joint collision probabilities for hundreds of rollouts against moving humans and rejects samples above a risk threshold, enabling smooth crowd navigation without freezing \cite{Trevisan2025}. Constrained Covariance-Steering MPPI augments the sampler with a low-level covariance-steering controller that shapes the state-distribution tube so the robot respects obstacle chance constraints with high probability \cite{Kouhestani2022}. Our Bayesian-Constraints MPPI follows this distribution-shaping philosophy but, unlike \cite{Trevisan2025}, needs no sample-rejection step—the probabilistic constraint model directly modulates the sampling law, allowing us to bound violation probability analytically.\\

\section{The Bayesian Constraints MPPI}
\begin{figure}[t]
  \centering

  \includegraphics[width=0.9\linewidth]{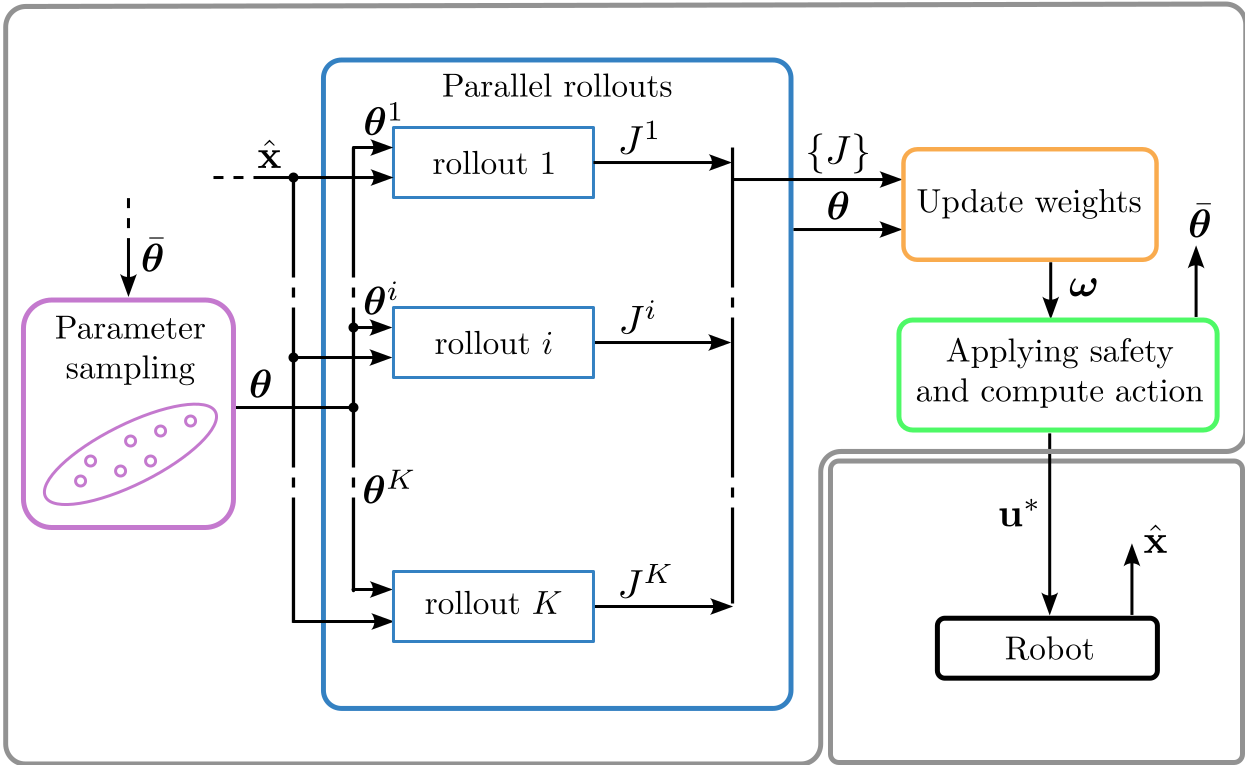}
\caption{BC\hbox{-}MPPI workflow.  A Gaussian sampler perturbs the
           nominal parameter vector $\bar{\boldsymbol{\theta}}$ and
           launches $K$ parallel roll-outs; their costs $J^{1},\dots,J^{K}$ feed the usual MPPI weight update,
           producing importance weights~$\omega$.  A Bayesian surrogate multiplies these weights by the joint feasibility probability, suppressing unsafe trajectories and yielding the filtered parameter $\boldsymbol{\theta}^{\!*}$.  The first input $\mathbf{u}^{\!*}$ of the associated control sequence is applied to the robot, closing the feedback loop with the
           measured state $\hat{\mathbf{x}}$.}
\label{fig:bc_mppi_overview}
\end{figure}
We now detail how a probabilistic safety layer can be fused with
Model-Predictive Path-Integral control.  The presentation proceeds
top-down: Secs.~\ref{subsec:mppi}–\ref{subsec:bayesian_constraints}
recap the standard MPPI sampler, describe the Bayesian surrogate
constraints, and show how the joint feasibility probability reshapes the sampling weights.  A high-level block diagram of the full pipeline is given in Fig.~\ref{fig:bc_mppi_overview}

\subsection{A Background on Model Predictive Path Integral}\label{subsec:mppi}

Model Predictive Path Integral (MPPI) control solves nonlinear, non-convex optimal–control problems by stochastic sampling rather than gradient search~\cite{Turrisi2024,Minarik2024}.  At each replanning step, it minimises  

\begin{equation}\label{eq:ocp}
\boldsymbol{\theta}^{\star}
  =\arg\!\min_{\boldsymbol{\theta}}\!
  \bigl[\ell_N(\mathbf{x}_N)+\!\!\sum_{i=0}^{N-1}\!\ell_i(\mathbf{x}_i,\mathbf{u}_i)\bigr]
  \;\text{s.t.}\;
  \mathbf{x}_{i+1}=\mathbf{f}(\mathbf{x}_i,\mathbf{u}_i),\;
  \mathbf{u}_i=\mathbf{\pi}(\boldsymbol{\theta},\mathbf{x}_i,t_i),\;
  \mathbf{x}_0=\hat{\mathbf{x}},
\end{equation}

where $\mathbf{\theta}$ is a low-dimensional parameter vector, $\mathbf{\pi}$ a (possibly state-dependent) input parametrisation—e.g.\ zero-order hold~\cite{williams2018}, cubic splines~\cite{Turrisi2024}, or Halton splines~\cite{Pezzato_2025}— $\hat{\mathbf{x}}$ the measured state, $\mathbf{f}$ the equation of motion of the dynamical system and $\ell_i$ the chosen cost function.

\paragraph{Sampling and weighting.}
We draw $K$ perturbations $\Delta\mathbf{\theta}^k\!\sim\!\mathcal N(0,\mathbf{\Sigma})$ about the previous mean $\bar{\mathbf{\theta}}$, roll out the dynamics, and evaluate
\begin{equation}
J^{k}:=\ell_N(\mathbf{x}_N^k)+\sum_{i=0}^{N-1}\ell_i(\mathbf{x}_i^k,\mathbf{u}_i^k).
\end{equation}

Each perturbation is weighted by an exponentiated cost
\begin{equation}\label{eq:weights}
\omega^k = \frac{\mu^k}{\sum_{j=1}^{K}\mu^j},
\quad
\mu^{k}= \exp\!\Bigl[-\tfrac{1}{\lambda}\bigl(J^{k}-\rho\bigr)\Bigr],
\end{equation}
yielding the updated parameter
\begin{equation}\label{eq:theta_variation}
\boldsymbol{\theta}^{\star}
  =\bar{\boldsymbol{\theta}}+\sum_{k=1}^{K}\omega^k\Delta\boldsymbol{\theta}^k,
\end{equation}
with temperature~$\lambda$ trading off exploration and exploitation~\cite{Minarik2024,rizzi_robust_2023}.

\paragraph{Execution.}
The first control in the optimised sequence is issued as
\begin{equation}\label{eq:input_parametrization}
\mathbf{u}^{\star}=\mathbf{\pi}(\boldsymbol{\theta}^{\star},\hat{\mathbf{x}},t_0),
\end{equation}
and held constant until the next replanning instant.  Because rollouts are embarrassingly parallel, modern MPPI implementations evaluate thousands of trajectories on a GPU in sub-millisecond time, enabling real-time control of agile UAVs and legged robots~\cite{Turrisi2024,Minarik2024}.

\subsection{Probabilistic Constraint Satisfaction in BC-MPPI}\label{subsec:bayesian_constraints}
In this subsection, we introduce the core mechanism that enforces
constraint satisfaction in \emph{Bayesian Constraints MPPI} (BC-MPPI).
The idea is directly inspired by the feasibility–weighted acquisition
rules of \emph{Constrained Bayesian Optimisation} (CBO)
\cite{Letham2019}: we reshape the sampling distribution so that samples
likely to violate any constraint receive exponentially smaller weight.\\

Let \(c_i(\boldsymbol{\theta})\le 0\) be the $i$-th constraint, modelled by a
surrogate model that returns a predictive mean~\(\mu_i(\boldsymbol{\theta})\) and standard deviation \(\sigma_i(\boldsymbol{\theta})\). Assuming additive Gaussian noise, the probability of feasibility is \(\Pr[c_i(\boldsymbol{\theta})\le 0]= \Phi\!\bigl(-\mu_i(\boldsymbol{\theta})/\sigma_i(\boldsymbol{\theta})\bigr)\), where \(\Phi\) is the normal CDF. Multiplying these probabilities across all \(I\) constraints yields the constraint–likelihood term \(\prod_{i=1}^{I}\Pr[c_i(\boldsymbol{\theta})\le 0]\). We attach that term to the original Gaussian sampling distribution \(\mathcal N(\boldsymbol{\theta}\mid\bar{\boldsymbol{\theta}},\boldsymbol{\Sigma})\),
obtaining the weighted density  

\begin{equation}\label{eq:constraint_weighted_prior}
p(\boldsymbol{\theta})
\;\;\propto\;\;
\mathcal N\!\bigl(\boldsymbol{\theta}\mid
           \bar{\boldsymbol{\theta}},\boldsymbol{\Sigma}\bigr)
\;
\prod_{j=1}^{J}\Pr\!\bigl[c_j(\boldsymbol{\theta})\le 0\bigr].
\end{equation}

Equation~\eqref{eq:constraint_weighted_prior} mirrors CBO’s strategy of multiplying an acquisition function by a joint feasibility probability \cite{Letham2019}.\\ 

Classic MPPI can be viewed as an importance–sampling estimator of the optimal control distribution.  At each iteration we draw i.i.d.\ perturbations \(\Delta\boldsymbol{\theta}^{k}\!\sim\!\mathcal N(\mathbf 0,\boldsymbol{\Sigma})\) around the current mean \(\bar{\mathbf{\theta}}\) and assign the importance weight
\begin{equation}
\mu^{k}= \exp\!\bigl[-(J^{k}-\rho)/\lambda\bigr],
\
\label{eq:Importance weight for classic MPPI}
\end{equation}

so that expectations under the \emph{target} density
\(
p^{\star}(\boldsymbol{\theta})\propto
\exp[-J(\boldsymbol{\theta})/\lambda]\,
\mathcal N(\boldsymbol{\theta}\mid\bar{\boldsymbol{\theta}},\boldsymbol{\Sigma})
\)
are approximated by a weighted average of samples from the \emph{sampling distribution} \(\mathcal N(\bar{\boldsymbol{\theta}},\boldsymbol{\Sigma})\).\\

BC-MPPI retains the same Gaussian sampling distribution but augments the target density with the joint feasibility probability,
\[
p^{\star}_{\!\text{BC}}(\boldsymbol{\theta})\propto
  \exp\!\bigl[-J(\boldsymbol{\theta})/\lambda\bigr]\,
  \mathcal N(\boldsymbol{\theta}\mid\bar{\boldsymbol{\theta}},\boldsymbol{\Sigma})\,
  \prod_{j=1}^{J}\Pr\!\bigl[c_{j}(\boldsymbol{\theta})\le 0\bigr].
\]
Because the sampling distribution is unchanged, the importance ratio acquires a single
extra factor, yielding the new weight
\[
\tilde{\mu}^{k}=
  \exp\!\bigl[-(J^{k}-\rho)/\lambda\bigr]\,
  \prod_{j=1}^{J}\Pr\!\bigl[c_{j}(\boldsymbol{\theta}^{k})\le 0\bigr].
\]

If the surrogate models predict that a sample \(\boldsymbol{\theta}^{k}\) will satisfy \emph{all} constraints with high probability, every factor \(\Pr[c_{j}(\boldsymbol{\theta}^{k})\le 0]\) is close to \(1\); the weight \(\tilde{\mu}^{k}\) therefore coincides with the classic MPPI weight
\(\mu^{k}\) and the sample contributes normally to the control update. Conversely, if any constraint is likely to be violated, at least one factor becomes vanishingly small, driving \(\tilde{\mu}^{k}\) towards zero. The sample is not discarded outright—it remains in the
Monte-Carlo pool—but its influence on the weighted average, and hence on the control law, is effectively null.  In this way BC-MPPI enforces constraints \emph{without} an explicit rejection step such as the one used in Dynamic Risk-Aware MPPI~\cite{Trevisan2025}, while preserving the
unbiased, gradient-free character of the original algorithm.

\section{Simulation Experiments}

\begin{figure}[t]
  \centering
  \begin{subfigure}{0.32\linewidth}
    \includegraphics[width=\linewidth]{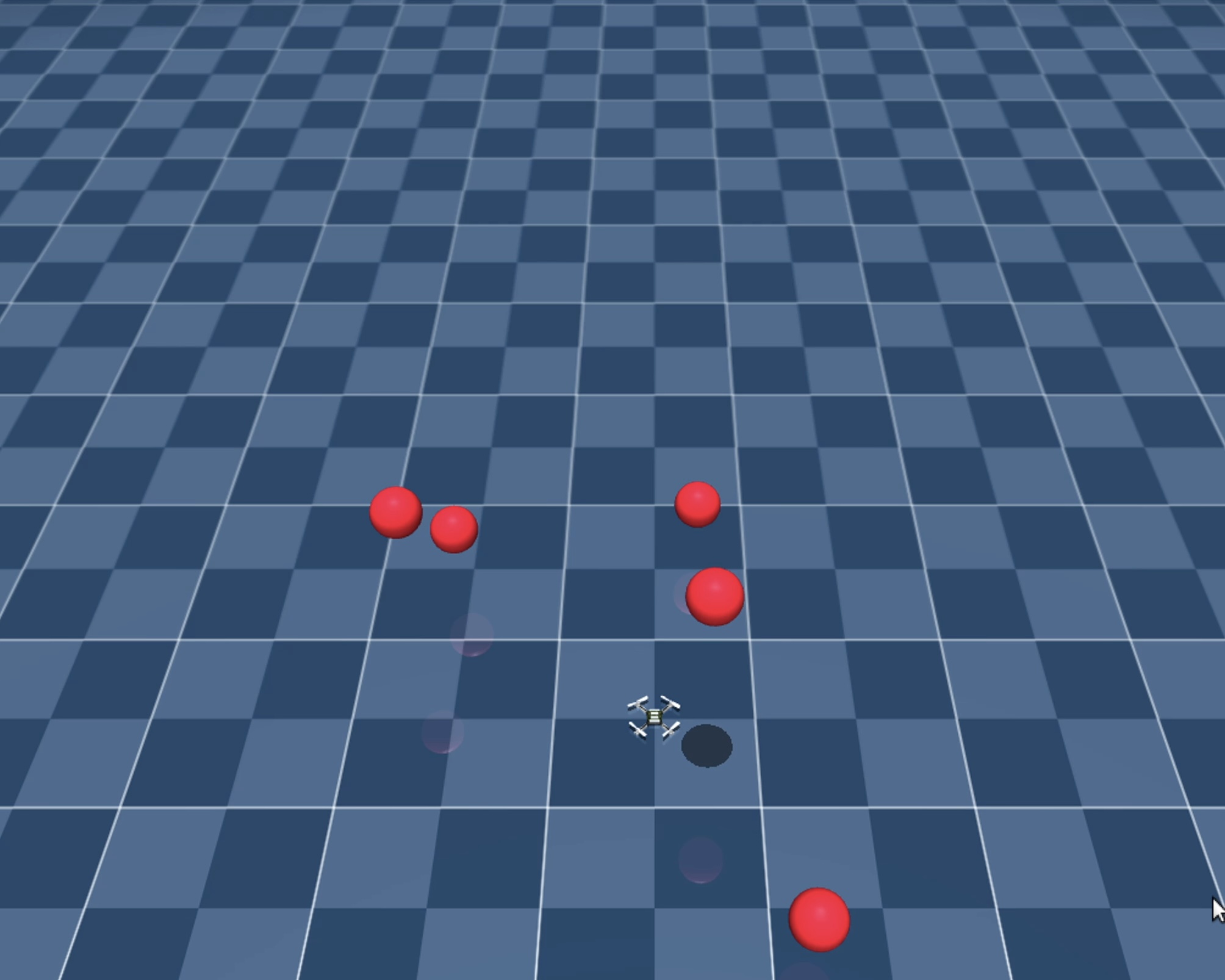}
  \end{subfigure}\hfill
  \begin{subfigure}{0.32\linewidth}
    \includegraphics[width=\linewidth]{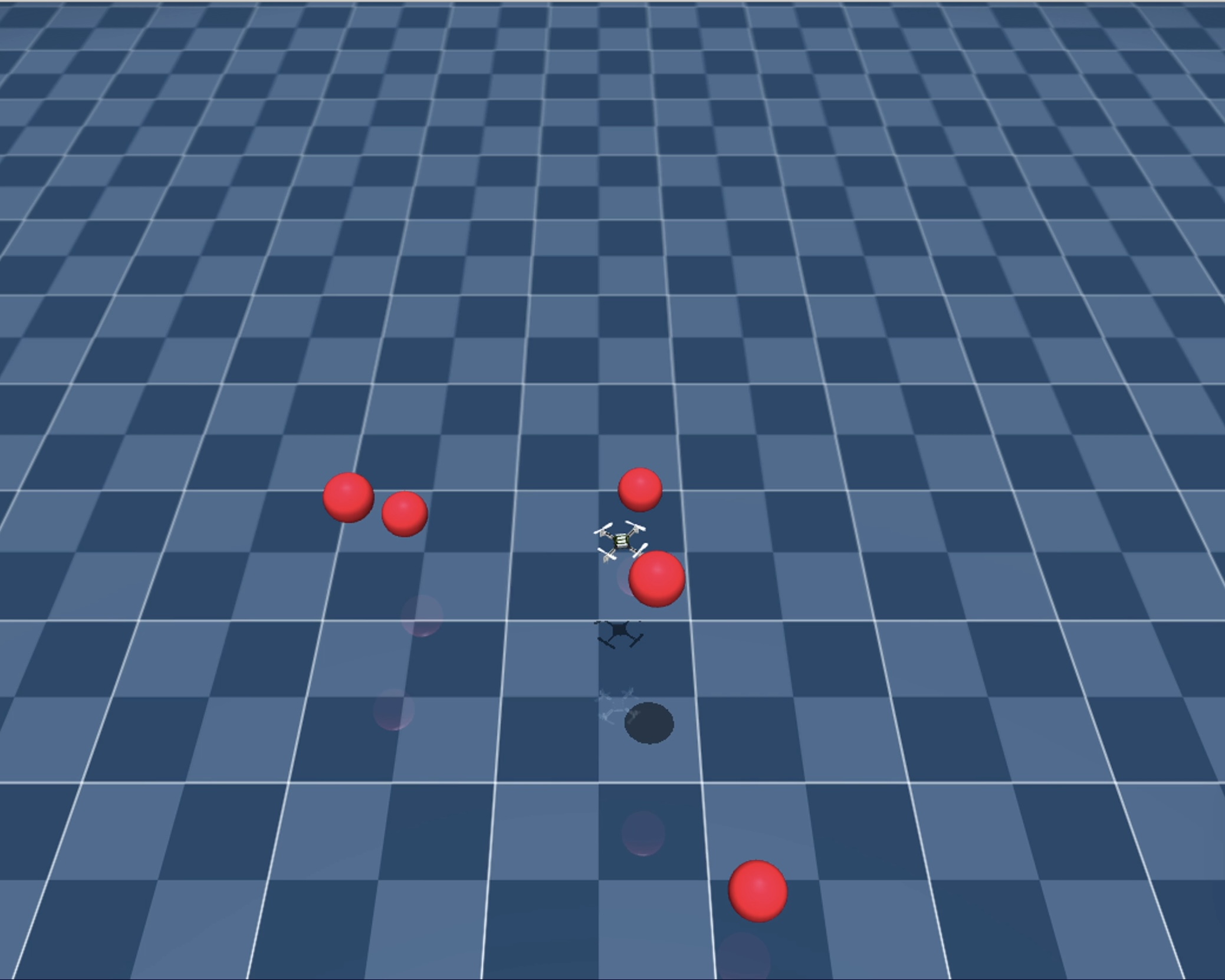}
  \end{subfigure}\hfill
  \begin{subfigure}{0.32\linewidth}
    \includegraphics[width=\linewidth]{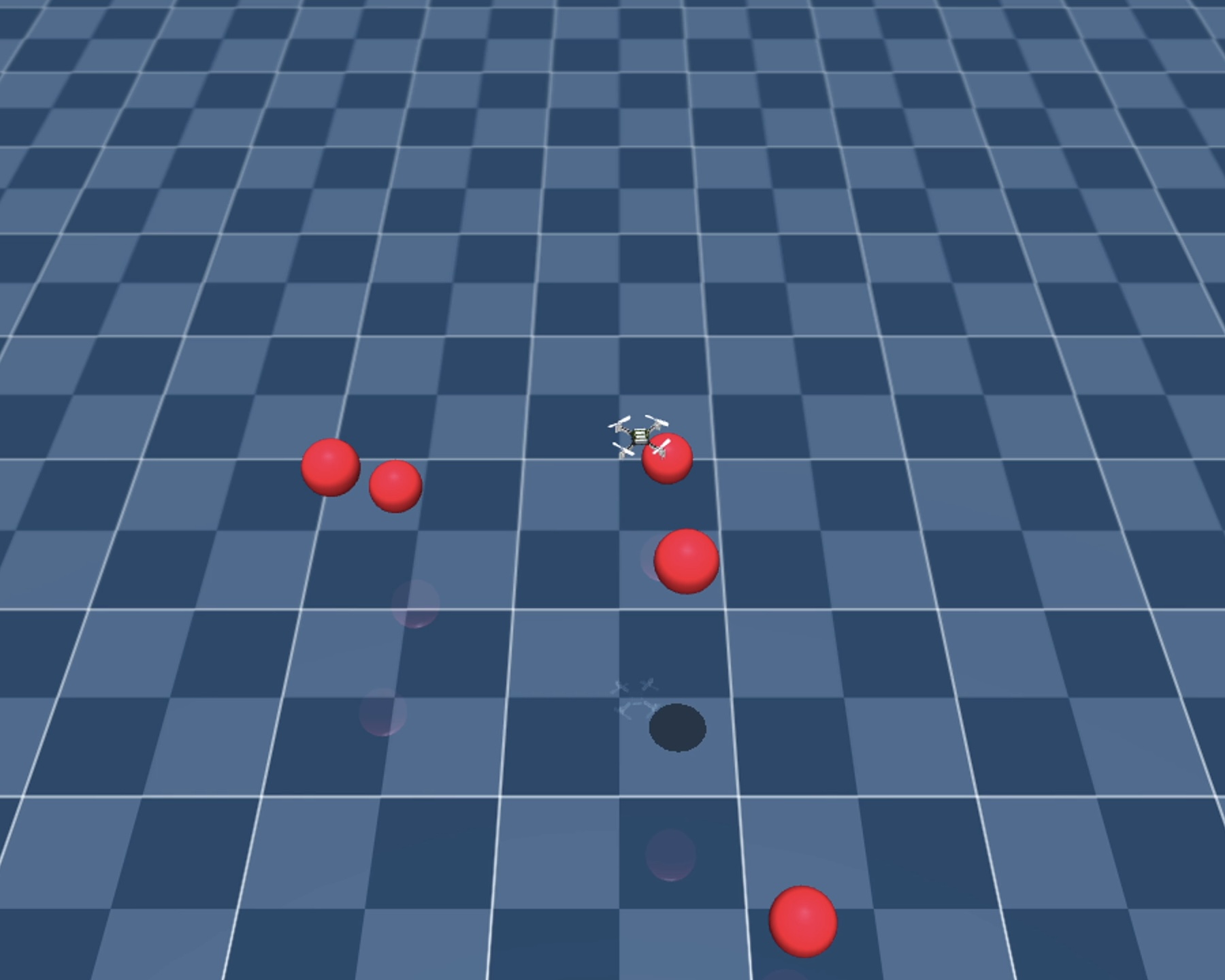}
  \end{subfigure}
  \vspace{-4pt}
  \caption{Snapshot sequence from a complex scenario involving five moving obstacles. Spheres model obstacles with inflated radii.}
  \label{fig:static_frames}
\end{figure}

 Classic MPPI, as described previously, applies an importance weighting to samples of the form in equation \eqref{eq:Importance weight for classic MPPI}, where samples with a higher cost are effectively rejected. Similarly, BC-MPPI – as described in equation \eqref{eq:constraint_weighted_prior} – applies an additional weighting layer to the samples based on their probability of constraint violation as predicted by the model, in our case a Bayesian Neural Network (BNN). To establish a baseline between these weighting methods, we also introduced a simplified version of MPPI – `MPPI-penalty', where constraint violations are handled only by applying a penalty term in the objective function, and there is no explicit rejection rule. Constraint violations are therefore accounted for by evaluating the following expression,

\begin{equation}
\text{constraint\_cost}_i =
\begin{cases} 
1e3, & \text{if } \text{constraint\_cost}_i < 0 \\ 
0, & \text{otherwise}
\end{cases}
\label{eq: cost function}
\end{equation}

where a constraint violation term \textit{D} is added to the cost based on the average L1 distance between the current position of the quadrotor and the obstacles, as described below.

\begin{equation}
D = (|x_2 - x_1|   + |y_2 - y_1| + |z_2 - z_1|) - r
\label{eq: l1 dist}
\end{equation}

We evaluated BC-MPPI in the MuJoCo physics engine \cite{todorov2012mujoco} on a quadrotor tasked with point-to-point flight in the presence of obstacles. Our method was evaluated against the two baselines: MPPI-penalty and Classic MPPI.\\

A wide range of scenarios was considered, including both stationary and moving obstacles. To systematically increase task complexity, the number of obstacles was increased, ranging from 3 to 15, and in the most demanding settings, both the obstacle trajectories and the target position were randomized at each iteration. These variations were designed to probe not only average-case performance but also the breakdown points of the respective methods.\\

For each scenario, we collected metrics characterizing both computational performance and trajectory quality. The performance metrics were simulation runtime - the duration of the simulation with each method – and the control frequency during the simulation. The metrics for trajectory quality were as follows: average distance of the quadrotor from the obstacles, total distance of the quadrotor from the target, and the number of collisions with obstacles that occurred. Additionally, for the rejection-based methods, we recorded the rejection rate, defined as the percentage of sampled rollouts that were discarded before evaluation.\\

Our final implementation of BC-MPPI employed a BNN surrogate. The BNN was trained on a dataset of 1000 simulated rollouts, where each sample included the initial system state (13 dimensions), the sequence of control inputs over a 25-step horizon (100 dimensions), and the corresponding constraint-violation term. The latter was computed as the average of the penalty terms across the horizon, giving a total input dimensionality of 113 and a single scalar output. To ensure data richness, rollouts were generated under three obstacle trajectory functions—circular, diagonal, and sinusoidal—combined in a 2:2:1 ratio. The dataset was shuffled, stored offline, and subsequently split into training and test sets in a 7:3 ratio. All features were standardized to mitigate scale bias and improve surrogate training.\\

On a held-out test set, the model achieved a mean squared error of 1.07 and an $R^{2}$ score of 0.08. Although the predictive accuracy could've been improved, the model proved sufficient for probabilistic weighting of rollouts in the closed-loop controller.\\

\subsection{BC-MPPI vs MPPI-penalty}

\begin{figure}[t]
  \centering
  \begin{subfigure}[b]{0.48\linewidth}
    \centering
    \includegraphics[width=\linewidth]{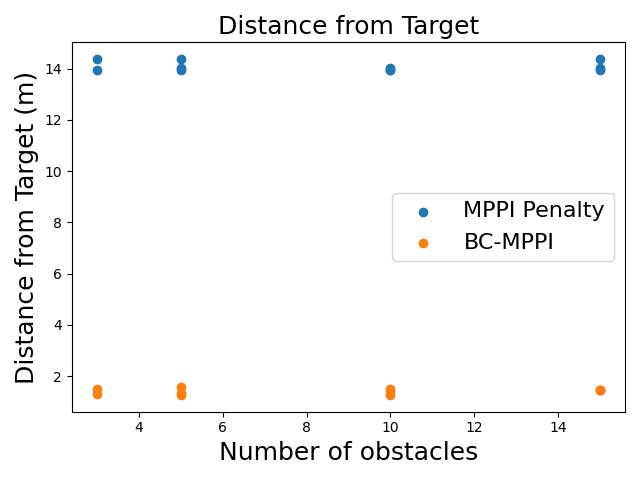}
    \label{fig:avg_dist}
  \end{subfigure}\hfill
  \caption{Distance of quadrotor from the target}
\end{figure}

Across all tested scenarios, MPPI-penalty failed to reach the target. This was consistent irrespective of the number or motion of obstacles. Consequently, the resulting trajectories exhibited both large average distances from the target and reduced distances from obstacles, as shown in Figure 3. The high obstacle proximity was particularly pronounced as the number of obstacles increased, leading to a significant rise in collisions. These observations reinforce that penalty-based shaping of the cost function is insufficient for enforcing safety in practice, as the optimizer continues to sample unsafe rollouts that are not filtered or strongly down-weighted.\\

From a computational perspective, MPPI-penalty achieved shorter simulation runtimes and higher control frequencies than BC-MPPI, especially in stationary-obstacle settings. However, the resulting large distances from obstacles and the target show that this computational efficiency came at the cost of safety: while control frequency was higher, the number of collisions grew rapidly as task complexity increased. This underscores the need for explicit rejection or probabilistic safety evaluation, since raw penalty terms proved unable to prevent unsafe rollouts from dominating the sampling distribution.

\subsection{BC-MPPI vs Classic MPPI}
When compared against Classic MPPI, BC-MPPI achieved clearer safety margins in the stationary-obstacle settings. As shown in Figure 3, the average distance from obstacles was consistently larger under BC-MPPI for low-to-moderate numbers of obstacles, confirming the surrogate’s ability to bias the sampling distribution toward safer rollouts. In more complex scenarios (moving obstacles with circular, diagonal and sinusoidal trajectories), this margin narrowed, with both methods yielding similar average obstacle clearances.\\

Interestingly, despite comparable distance from obstacles in complex cases, BC-MPPI consistently tracked the target more closely (Fig. 4). While the absolute difference in average distance to the target was small (on the order of centimeters), the effect was consistent across trials. In safety-critical applications, where repeated small deviations can accumulate into large errors over time, such improvements remain practically significant.\\

Most importantly, in the most complex scenarios, BC-MPPI produced fewer collisions than Classic MPPI while also maintaining a lower rejection rate overall (Fig. 6). This indicates that BC-MPPI uses samples more efficiently: unsafe rollouts are not discarded outright, but are probabilistically down-weighted, preserving useful gradient-free exploration while still improving safety.\\

\begin{figure}[t]
  \centering
  \begin{subfigure}[b]{0.48\linewidth}
    \centering
    \includegraphics[width=\linewidth]{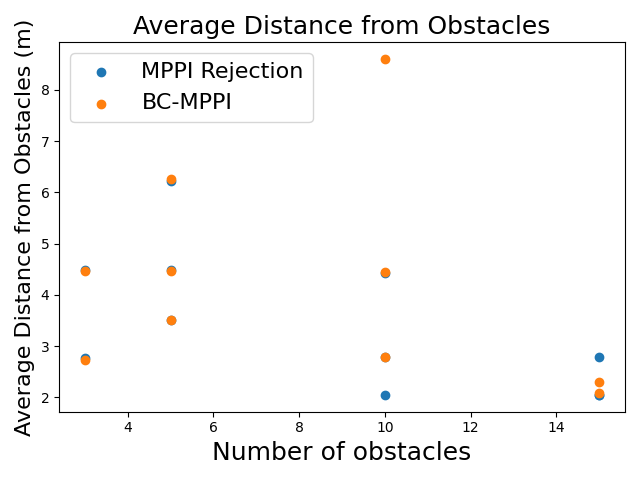}
    \caption{Moving obstacle scenario}
    \label{fig:avg_dist}
  \end{subfigure}\hfill
  \begin{subfigure}[b]{0.48\linewidth}
    \centering
    \includegraphics[width=\linewidth]{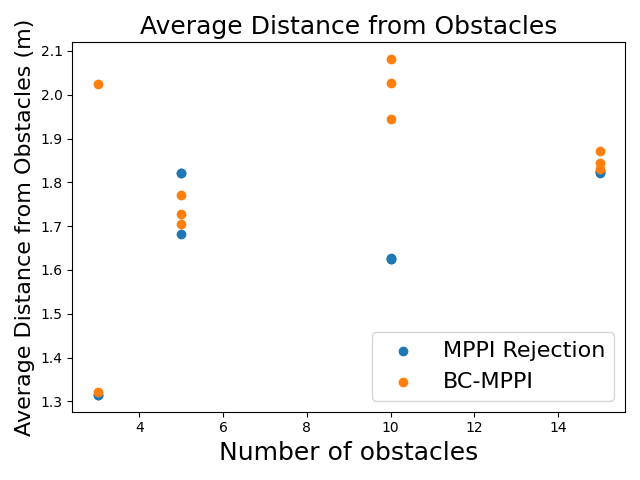}
    \caption{Stationary obstacle scenario}
    \label{fig:avg_dist_stationary}
  \end{subfigure}
  \caption{Average obstacle distance of BC-MPPI vs Classic MPPI in scenarios with stationary and moving obstacles.}
 \label{fig:static_metrics}
\end{figure}

\begin{figure}[t]
  \centering
  \begin{subfigure}[b]{0.48\linewidth}
    \centering
    \includegraphics[width=\linewidth]{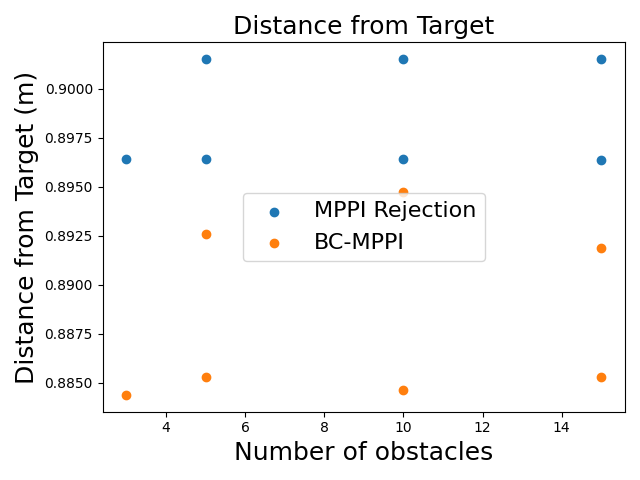}
    \label{fig:avg_dist}
  \end{subfigure}\hfill
  \caption{Distance of quadrotor from the target}
\end{figure}

\begin{figure}[t]
  \centering
  \begin{subfigure}[b]{0.48\linewidth}
    \centering
    \includegraphics[width=\linewidth]{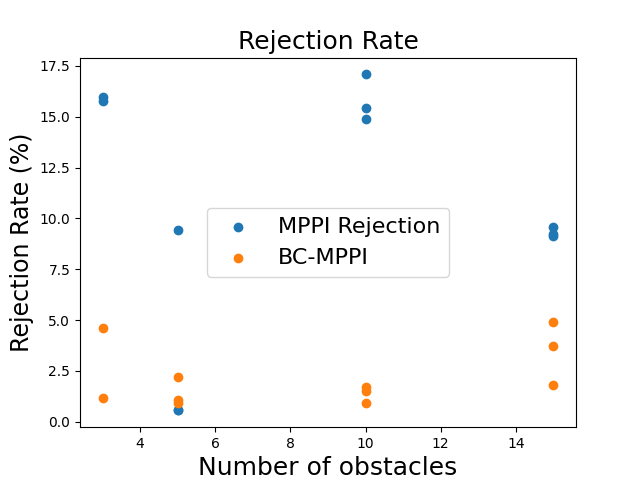}
    \label{fig:avg_dist}
  \end{subfigure}\hfill
  \caption{Rejection rate}
\end{figure}

Finally, Classic MPPI demonstrated higher control frequencies across nearly all scenarios, making it more responsive to dynamic changes in obstacle trajectories. This responsiveness, however, comes at the expense of a higher rejection rate and greater variability in trajectory quality, whereas BC-MPPI traded computational efficiency for more consistent safety and goal-tracking.\\

\section{Discussion}

The Bayesian–Constraints layer converts MPPI—an inherently stochastic, gradient–free controller—into a closed–loop scheme with an explicit, probabilistic notion of safety. Two properties make the approach amenable to Verification and Validation (V\&V). \emph{First}, the BNN surrogate is learned offline from a finite, version–controlled dataset; its predictive mean and variance form a static artefact that can be unit–tested, regression–tested as new data arrive, or subjected to formal probabilistic checks (e.g., proving that the posterior variance never exceeds a threshold over the reachable state space). \emph{Second}, at runtime every trajectory receives a single scalar weight
\begin{equation*}
w(\boldsymbol{\theta}) = \prod_{j}\Pr\big[c_j(\boldsymbol{\theta}) \le 0\big],
\end{equation*}
which monotonically reflects constraint satisfaction. This scalar is easily monitored by a lightweight runtime guard: if $w(\theta)$ falls below a certified bound, control can be handed over to a simpler, formally verified safe mode.\\

The experimental results further highlight how this architecture balances computational cost with safety guarantees. Compared to MPPI–penalty, BC–MPPI was the only method that successfully reached the target across all scenarios, whereas penalty–based control failed even in simple cases, confirming that penalty shaping alone is inadequate for enforcing hard safety requirements. Against classic MPPI, BC–MPPI achieved comparable or better obstacle clearance and consistently maintained closer target tracking, while also reducing collisions in the most complex environments. Importantly, these improvements were obtained with a lower rejection rate, showing that probabilistic weighting can use samples more efficiently than hard rejection.\\

A key trade–off is computational: BC–MPPI incurred longer simulation runtimes and lower control frequencies due to surrogate evaluation and weighting. Although the baselines were more responsive, their higher collision rates and reduced reliability illustrate that raw frequency is not synonymous with safety. For safety–critical applications, the modest loss in control rate is offset by a measurable reduction in constraint violations and the ability to certify probabilistic guarantees.\\

The architecture therefore yields a clean separation of concerns: low–level dynamics remain in fast, handwritten code; the surrogate and weighting rule reside in a self–contained module that can be verified with traditional software–engineering practices (code review, static analysis, continuous integration); and high–level mission logic can rely on the weight as a measurable contract, simplifying compositional reasoning about the overall system.\\

\section{Limitations and Future Work}

 While the present results demonstrate the feasibility of BC-MPPI, several limitations remain. First, the surrogate accuracy was modest $R^{2} \approx 0.08$, yet still sufficient for weighting; more expressive models or larger datasets may improve predictive fidelity without compromising runtime. Second, the experiments focused on a quadrotor in obstacle-avoidance tasks; scaling to higher-dimensional robots or more diverse environments may reveal new challenges in training efficiency and model generalization. Finally, although the additional runtime cost was acceptable in our setting, deploying BC-MPPI on embedded hardware will require further optimization of surrogate inference and sampling. Addressing these limitations—through model compression, online updating of the surrogate, or integration with other safety-filtering methods—offers a promising direction for future work.

\section{Conclusion}
We have shown that embedding a probabilistic constraint layer into MPPI yields a controller that is both agile and safety-aware, outperforming a classic penalty-based baseline in static and dynamic obstacle fields. Future work will develop automatic coverage metrics for the offline data set, perform formal co-analysis of the surrogate model with temporal-logic mission specifications, and design incremental re-training pipelines that preserve previously proven safety margins while adapting to changing environments.

%
%
%
\bibliographystyle{splncs04}
\bibliography{mybibliography}

\begin{thebibliography}{10}
\providecommand{\url}[1]{\texttt{#1}}
\providecommand{\urlprefix}{URL }
\providecommand{\doi}[1]{https://doi.org/#1}

\bibitem{belvedere2025feedbackmppi}
Belvedere, T., Ziegltrum, M., Turrisi, G., Modugno, V.: Feedback-{MPPI}: Fast sampling-based mpc via rollout differentiation -- adios low-level controllers. arXiv preprint 2506.14855  (2025)

\bibitem{Carius2022}
Carius, J., Ranftl, R., Farshidian, F., Hutter, M.: Constrained stochastic optimal control with learned importance sampling: A path integral approach. The International Journal of Robotics Research  \textbf{41}(2),  189--209 (2022)

\bibitem{Gandhi2021}
Gandhi, M.S., Vlahov, B., Gibson, J., Williams, G., Theodorou, E.A.: Robust model predictive path integral control: Analysis and performance guarantees. IEEE Robotics and Automation Letters  \textbf{6}(2),  3653--3660 (2021)

\bibitem{howell2022}
Howell, T., Gileadi, N., Tunyasuvunakool, S., Zakka, K., Erez, T., Tassa, Y.: Predictive {Sampling}: {Real}-time {Behaviour} {Synthesis} with {MuJoCo}, arXiv:2212.00541

\bibitem{Katayama2023}
Katayama, S., Murooka, M., Tazaki, Y.: Model predictive control of legged and humanoid robots: models and algorithms. Advanced Robotics  \textbf{37}(5),  298--315 (2023)

\bibitem{Kicki2025}
Kicki, P.: Low-pass sampling in model predictive path integral control. arXiv preprint arXiv:2503.11717  (2025)

\bibitem{Kobilarov2012}
Kobilarov, M.: Cross-entropy motion planning. The International Journal of Robotics Research  \textbf{31}(7),  855--871 (2012)

\bibitem{Kouhestani2022}
Kouhestani, A., Bakolas, E., Theodorou, E.A.: Constrained covariance steering model predictive path integral control. In: Proceedings of the IEEE Conference on Decision and Control. pp. 3281--3288 (2022)

\bibitem{Letham2019}
Letham, B., Karrer, B., Ottoni, G., Bakshy, E.: Constrained bayesian optimization with noisy experiments. Bayesian Analysis  \textbf{14}(2),  495--519 (2019)

\bibitem{Liu2021}
Liu, Z., Zhou, H., Chen, B., Zhong, S., Hebert, M., Zhao, D.: Constrained model-based reinforcement learning with robust cross-entropy method. arXiv:2010.07968 (2021)

\bibitem{Minarik2024}
Minařík, M., Pěnička, R., Vonásek, V., Saska, M.: Model predictive path integral control for agile unmanned aerial vehicles. In: 2024 IEEE/RSJ International Conference on Intelligent Robots and Systems (IROS). pp. 13144--13151 (2024)

\bibitem{Neal1995}
Neal, R.M.: Bayesian learning for neural networks. Ph.D. thesis, CAN (1995)

\bibitem{Parwana2024}
Parwana, H., Black, M., Hoxha, B., Okamoto, H., Fainekos, G., Prokhorov, D., Panagou, D.: Risk-aware {MPPI} for stochastic hybrid systems. arXiv:2411.09198 (2024)

\bibitem{Pezzato_2025}
Pezzato, C., Salmi, C., Trevisan, E., Spahn, M., Alonso-Mora, J., Hernández~Corbato, C.: Sampling-based model predictive control leveraging parallelizable physics simulations. IEEE Robotics and Automation Letters  \textbf{10}(3),  2750--2757 (2025)

\bibitem{Qin2003}
Qin, S., Badgwell, T.A.: A survey of industrial model predictive control technology. Control Engineering Practice  \textbf{11}(7),  733--764 (2003)

\bibitem{Rasmussen2006}
Rasmussen, C.E., Williams, C.K.I.: Gaussian Processes for Machine Learning (Adaptive Computation and Machine Learning). The MIT Press (2005)

\bibitem{rizzi_robust_2023}
Rizzi, G., Chung, J.J., Gawel, A., Ott, L., Tognon, M., Siegwart, R.: Robust {Sampling}-{Based} {Control} of {Mobile} {Manipulators} for {Interaction} {With} {Articulated} {Objects}. IEEE Transactions on Robotics  \textbf{39}(3),  1929--1946 (Jun 2023)

\bibitem{todorov2012mujoco}
Todorov, E., Erez, T., Tassa, Y.: Mujoco: A physics engine for model-based control. In: 2012 IEEE/RSJ International Conference on Intelligent Robots and Systems. pp. 5026--5033. IEEE (2012)

\bibitem{Trevisan2025}
Trevisan, E., Mustafa, K.A., Notten, G., Wang, X., Alonso-Mora, J.: Dynamic risk-aware {MPPI} for mobile robots in crowds via efficient monte carlo approximations. arXiv:2506.21205 (2025), accepted at IROS~2025

\bibitem{Turrisi2024}
Turrisi, G., Modugno, V., Amatucci, L., Kanoulas, D., Semini, C.: On the benefits of gpu sample-based stochastic predictive controllers for legged locomotion. In: 2024 IEEE/RSJ International Conference on Intelligent Robots and Systems (IROS). pp. 13757--13764 (2024)

\bibitem{Wensing2024}
Wensing, P.M., Posa, M., Hu, Y., Escande, A., Mansard, N., Prete, A.D.: Optimization-based control for dynamic legged robots. IEEE Transactions on Robotics  \textbf{40},  43--63 (2024)

\bibitem{Williams2017}
Williams, G., Aldrich, A., Theodorou, E.A.: Model predictive path integral control: From theory to parallel computation. Journal of Guidance, Control, and Dynamics  \textbf{40}(2),  344--357 (2017)

\bibitem{Williams2016}
Williams, G., Drews, P., Goldfain, B., Rehg, J.M., Theodorou, E.A.: Aggressive driving with model predictive path integral control. In: Proceedings of the IEEE International Conference on Robotics and Automation (ICRA). pp. 1433--1440 (2016)

\bibitem{williams2018}
Williams, G., Drews, P., Goldfain, B., Rehg, J.M., Theodorou, E.A.: Information-{Theoretic} {Model} {Predictive} {Control}: {Theory} and {Applications} to {Autonomous} {Driving}. IEEE Transactions on Robotics  \textbf{34}(6),  1603--1622 (Dec 2018)

\bibitem{Yin2023}
Yin, J., Dawson, C., Fan, C., Tsiotras, P.: Shield model predictive path integral: A computationally efficient robust mpc method using control barrier functions. IEEE Robotics and Automation Letters  \textbf{8}(11),  7106--7113 (2023)

\bibitem{Yin2022}
Yin, J., Zhang, Z., Tsiotras, P.: Risk-aware model predictive path integral control using conditional value-at-risk. In: 2023 IEEE International Conference on Robotics and Automation (ICRA). pp. 7937--7943 (2023)

\end{thebibliography}

\end{document}